\def\year{2020}\relax
\documentclass[letterpaper]{article} 
\usepackage{aaai20}  
\usepackage{times}  
\usepackage{helvet} 
\usepackage{courier}  
\usepackage[hyphens]{url}  
\usepackage{graphicx} 
\usepackage{graphicx}  
\usepackage{latexsym}
\usepackage{booktabs}
\usepackage{amsmath, amssymb}
\usepackage{bm}
\usepackage{caption}
\usepackage{subfig}
\usepackage{multirow}
\usepackage{algorithm2e}
\usepackage{enumitem}
\usepackage{fancyvrb}
\usepackage{relsize}
\usepackage{tipa}
\usepackage{tikz}
\usepackage{pgfplots}

\DeclareMathOperator*{\minimize}{minimize}
\newcommand\restr[2]{{
  \left.\kern-\nulldelimiterspace 
  #1 
  \vphantom{\big|} 
  \right|_{#2} 
  }}

\title{Towards Zero-shot Learning for Automatic Phonemic Transcription}

\author{Xinjian Li, Siddharth Dalmia, David R. Mortensen, Juncheng Li, Alan W Black, Florian Metze\\ 
  Language Technologies Institute, School of Computer Science\\
  Carnegie Mellon University \\
  {\tt \{xinjianl, sdalmia, dmortens, junchenl, awb, fmetze\}@cs.cmu.edu} 
}

\begin{document}

\maketitle

\begin{abstract}
Automatic phonemic transcription tools are useful for low-resource language documentation. However, due to the lack of training sets, only a tiny fraction of languages have phonemic transcription tools. Fortunately, multilingual acoustic modeling provides a solution given limited audio training data. A  more challenging problem is to build phonemic transcribers for languages with zero training data. The difficulty of this task is that phoneme inventories often differ between the training languages and the target language, making it infeasible to recognize unseen phonemes. In this work, we address this problem by adopting the idea of zero-shot learning. Our model is able to recognize unseen phonemes in the target language without any training data. In our model, we decompose phonemes into corresponding articulatory attributes such as \textit{vowel} and \textit{consonant}. Instead of predicting phonemes directly, we first predict distributions over articulatory attributes, and then compute phoneme distributions with a customized acoustic model. We evaluate our model by training it using 13 languages and testing it using 7 unseen languages. We find that it achieves 7.7\% better phoneme error rate on average over a standard multilingual model.
\end{abstract}

\section{Introduction}
Over the last decade, automatic speech recognition (ASR) has achieved great successes in many rich-resourced languages such as English, French and Mandarin. On the other hand, speech resources are still sparse for the majority of other languages. They cannot thus benefit directly from recent technologies. As a result, there is an increasing interest in building speech processing systems for low-resource languages. In particular, phoneme transcription tools  are useful for low-resource language documentation by improving workflow for linguists to analyze those languages~\cite{adams2018evaluating,michaud2018integrating}.



A more challenging task is to transcribe phonemes in the language with zero training data. This task has significant implications in documenting endangered languages and preserving the associated cultures~\cite{gippert2006essentials}. This data setup has mainly been studied in the unsupervised speech processing field~\cite{glass2012towards,versteegh2015zero,hermann2018multilingual}, which typically uses an unsupervised technique to learn representations which can be used towards speech processing tasks. 

However, those unsupervised approaches could not generate phonemes directly and there has been few works studying zero-shot learning for unseen phonemes transcription, which consist of learning an acoustic model without any audio data or text data for a given target language and unseen phonemes. In this work, we aim to solve this problem to transcribe unseen phonemes for unseen languages without considering any target data, audio or text.


\begin{figure*}[t]
  \centering
  \includegraphics[width=0.8\textwidth]{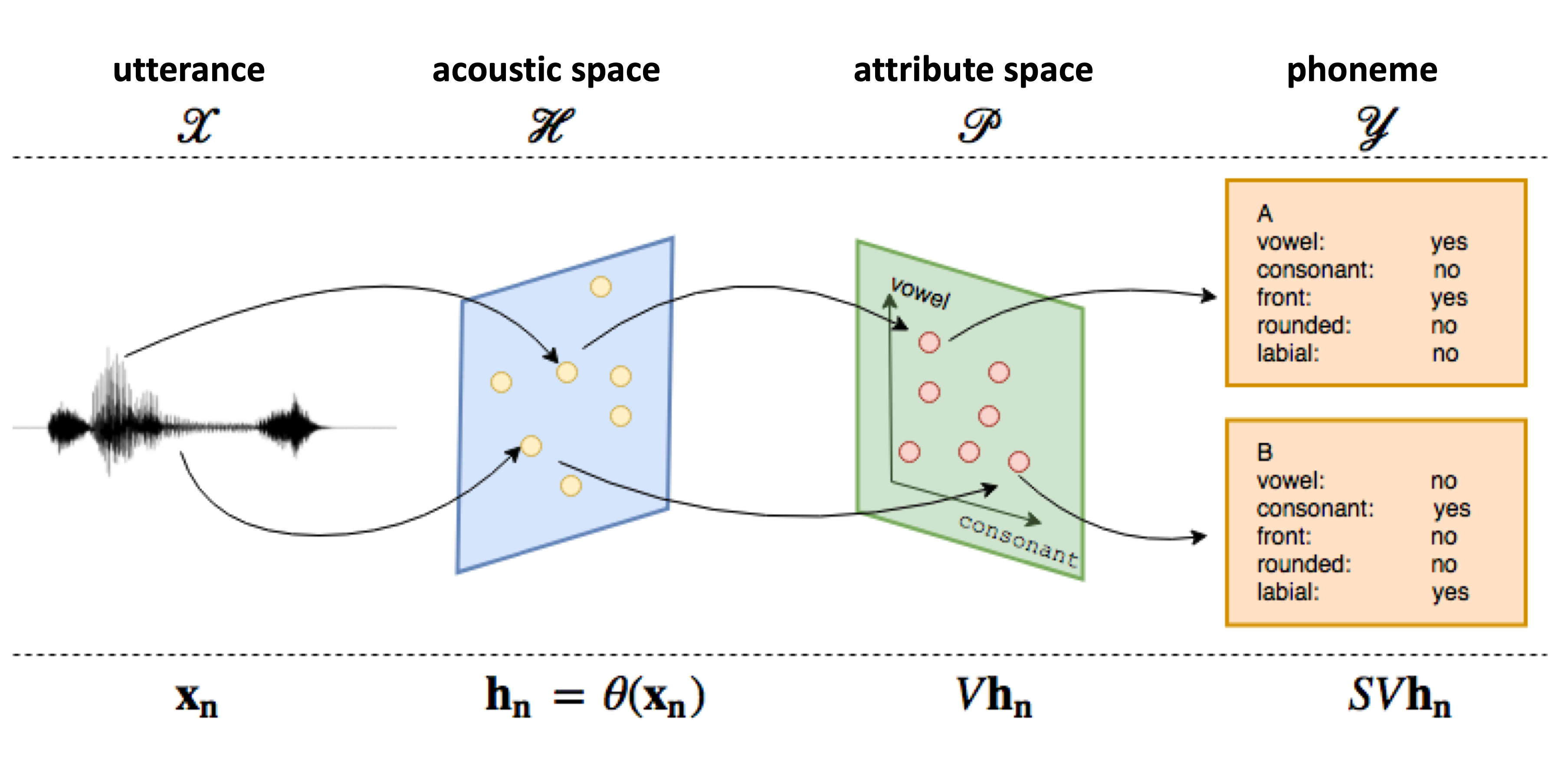}
  \caption{Illustration of the proposed zero-shot learning framework. Each utterance is first mapped into acoustic space (or hidden space) $\mathcal{H}$. Then we transform each point in the acoustic space into attribute space $\mathcal{P}$ with a linear transformation $V$. Finally phoneme distributions can be obtained by applying a signature matrix $S$}
  \label{fig:arch}
\end{figure*}

The prediction of unseen objects has been studied for a long time in the computer vision field. For specific object classes such as \textit{faces}, \textit{vehicles} and \textit{cats}, a significant number manually labeled data is usually available, but collecting sufficient data for every object human could recognize is impossible. Zero-shot learning attempts to solve this problem to classify unseen objects using mid-level side information. For example, \textit{zebra} can be recognized by detecting attributes such as \textit{stripped}, \textit{black} and \textit{white}. Inspired by approaches in computer vision research, we propose the Universal Phonemic Model (UPM) to apply zero-shot learning to acoustic modeling. In this model, we decompose the phoneme into its attributes and learn to predict a distribution over various articulatory attributes. For example, the phoneme /a/ can be decomposed into its attributes: \textit{vowel}, \textit{open}, \textit{front} and \textit{unrounded}. This can then be used to infer the unseen phonemes for the test language as the unseen phonemes can be decomposed into common attributes covered in the training phonemes.


Our approach is summarized in Figure~\ref{fig:arch}. First, frames are extracted and a standard acoustic model is applied to map each frame into the acoustic space (or hidden space) $\mathcal{H}$. Next we transform it into the attribute space $\mathcal{P}$ which reflects the articulatory distribution of each frame (such as whether it indicates a \textit{vowel} or a \textit{consonant}). Then, we compute the distribution of phonemes for that frame using a predefined signature matrix $S$ which describes relationships between articulatory attributes and phonemes in each language. 

To evaluate our UPM approach, we trained the model on 13 languages and tested it on another 7 languages. We also trained a multilingual acoustic model as a baseline for comparison. The result indicates that we consistently outperform the baseline multilingual model, and we achieve 7.7\% improvements in phoneme error rate on average.

The main contributions of this paper are as the followings:
\begin{enumerate}
    \item We propose the Universal Phonemic Model (UPM) that can recognize unseen phonemes during training by incorporating knowledge from the phonetics/phonology domain.
    \item We introduce a sequence prediction model to integrate a zero-shot learning framework for sequence prediction problem. 
    \item We show that our model is effective for 7 different languages, and our model gets 7.7\% better phoneme error rate over the baseline on average.
\end{enumerate}

\section{Approach}

\begin{figure*}[t]
  \centering
  \includegraphics[width=0.85\textwidth]{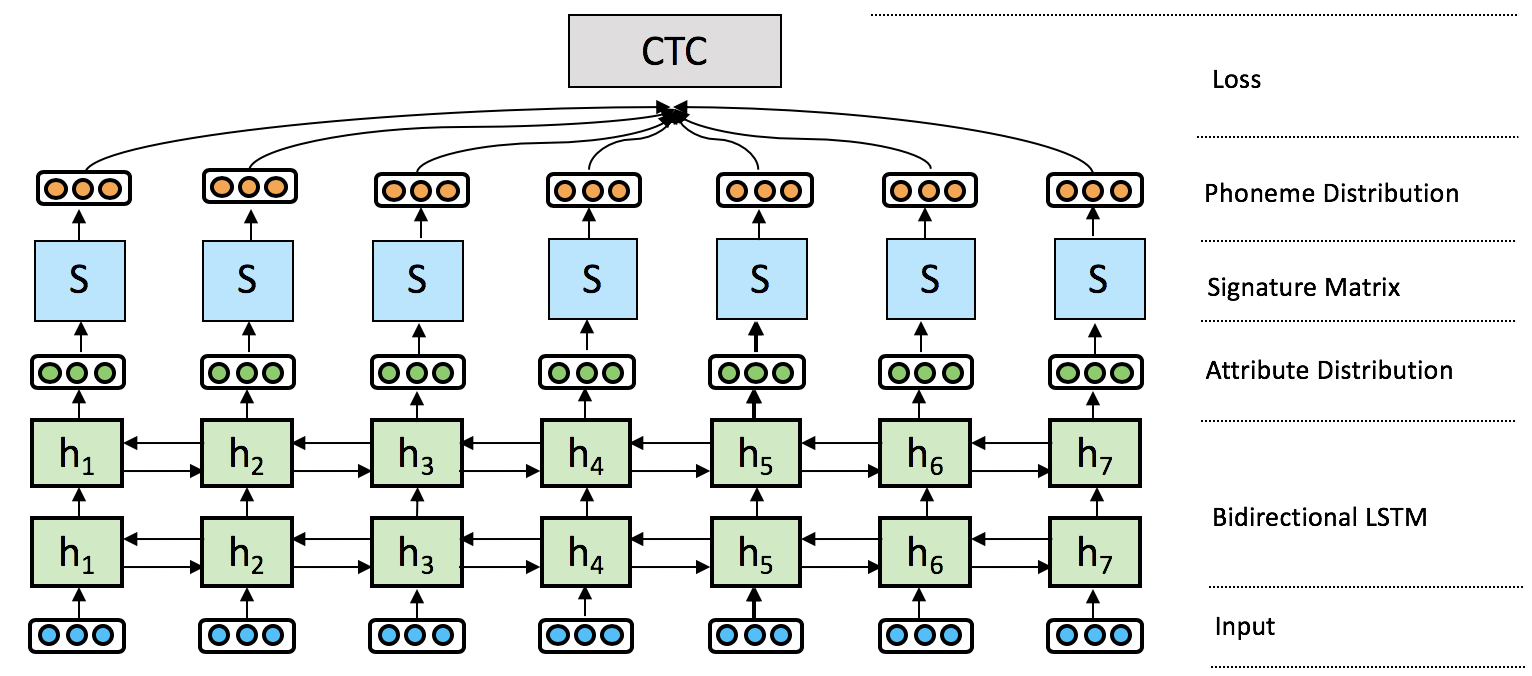}
  \caption{Illustration of the sequence model for zero-shot learning. The input layer is first processed with a Bidirectional LSTM acoustic model, and produces a distribution over articulatory attributes. Then it is transformed into a phoneme distribution by a language dependent signature matrix $S$}
  \label{fig:model}
\end{figure*}

This section explains the details of our Universal Phonemic Model (UPM). In the first section, we describe how we constructed a proper set of articulatory attributes for acoustic modeling. Next, we demonstrate how to assign attributes to each phoneme by giving an algorithm to parse X-SAMPA format. Finally we show how we integrate the phonetic information into the sequence model with a CTC loss~\cite{graves2006connectionist}.
\subsection{Articulatory Attributes}
Unlike attributes in the computer vision field, attributes of phonemes are independent of the corpus and dataset, they are well investigated and defined in the domain of articulatory phonetics~\cite{ladefoged2014course}. Articulatory phonetics describes the mechanism of speech production such as the manner of articulation and placement of articulation, and it tends to describe phones using discrete features such as voiced, bilabial (made with the two lips) and fricative. These articulatory features have been shown to be useful in speech recognition~\cite{kirchhoff1998combining,stuker2003multilingual,muller2017improving}, and are a good choice for attributes for our purpose. We provide some categories of articulatory attributes below. 
\\\textbf{Consonants}. Consonants are formed by obstructing the airstream through the vocal tract. They can be categorized in terms of the placement and the manner of this obstruction. The placements can be largely divided into three classes: \textit{labial}, \textit{coronal}, \textit{dorsal}. Each of the class have more fine-grained classes.  The manners of articulation can be grouped into: \textit{stop}, \textit{fricative}, \textit{approximant} etc.
\\\textbf{Vowel}. In the production of vowels, the airstream is relatively unobstructed. Each vowel sound can be specified by the positions of lips and tongue~\cite{ladefoged2014course}. For instance, the tongue is at its highest point in the front of the mouth for \textit{front} vowels. Additionally, vowels can be characterized by properties such as whether the lips are rounding or not (\textit{rounded}, \textit{unrounded}).
\\\textbf{Diacritics}. Diacritics are small marks to modify vowels and consonants by attaching to them. For instance, \textit{nasalization} marks a sound for which the velopharyngeal port is open and air can pass through the nose. To make the articulatory attribute set manageable, we assign attributes of diacritics to some existing consonants attributes if they share similar articulatory property. For example, \textit{nasalization} is treated as the \textit{nasal} attribute in consonants.

In addition to articulatory attributes mentioned above, we note that we also need to allocate an special attribute for blank in order to predict blank labels in CTC model, and backpropagate their gradients into the acoustic model. Thus, our articulatory attribute set $A_{phone}$ is defined as the union of these three domain attributes as well as the blank label, 
\begin{align*}
    A_{phone} = A_{consonants} &\cup A_{vowels} \\
    &\cup A_{diacritics} \cup \bigl\{ blank \bigr\}
\end{align*}

\subsubsection{Attribute Assignment}

Next, we need to assign each phoneme with appropriate attributes. There are multiple approaches to retrieve articulatory attributes. The simplest one is to use tools to collect articulatory features for each phoneme \cite{mortensen2016panphon}. However, those tools only provide coarse-grained phonological features but we expect more fine-grained and customized articulatory features. In this section, we propose a naive but useful approach for attribute assignment. We note that we use X-SAMPA format to denote each IPA in this work. X-SAMPA was devised to produce a computer-readable representation for IPA. Each IPA segment can be mapped to X-SAMPA with appropriate rule-based tools~\cite{mortensen2018epitran}. For example, IPA /\textipa{@}/ can be represented as /@/ in X-SAMPA. 

\begin{algorithm}
\SetAlgoLined
\SetKwInOut{Input}{input}\SetKwInOut{Output}{output}
 \Input{X-SAMPA representation of phoneme $p$}
 \Output{Articulatory attribute set $A \subseteq A_{phone}$ for $p$ }
 \BlankLine
 $A \leftarrow$ empty set  \;
 \BlankLine
 \While{$p \not\in P_{base}$}{
  find the longest suffix $p_{s} \in P_{base}$ \;
  
  Add $\restr{f}{P_{base}}(p_{s})$ to $A$ \;
  
  Remove suffix $p_{s}$ from $p$ \;
 }
 Add $\restr{f}{P_{base}}(p)$ to $A$ \
 \caption{A simple algorithm to assign attributes to phonemes}
 \label{algo:attribute}
\end{algorithm}


The assignment can be formulated as the problem to construct an assignment function $f: P_{xsampa} \rightarrow 2^{A_{phone}}$ where the domain $P_{xsampa}$ is the set of all valid X-SAMPA phonemes, and the range $2^{A_{phone}}$ is a subset of articulatory attributes for each phoneme . The assignment function should map each phoneme into its corresponding subset of $A_{phone}$. To construct the function in the entire domain $P_{xsampa}$, we first manually map a small subset $P_{base} \subset{P_{xsampa}} $ and construct a restricted assignment function $\restr{f}{P_{base}}:  P_{base} \rightarrow 2^{A_{phone}}$. The mapping is customizable and has been verified with the IPA handbook \cite{decker1999handbook}. Then for every phoneme $p \in P_{xsampa}$, we continue to remove diacritics suffix from it until it could be found in $P_{base}$. For example, to recognize /ts\textunderscore\textgreater/, we can first match the suffix, /\textunderscore\textgreater/ as an \textit{ejective}, and then recognize /ts/ as a consonant defined in $P_{base}$. The Algorithm~\ref{algo:attribute} summarizes our approach. 
  
  

\subsection{Sequence model for zero-shot learning}

\begin{table*}[t]
\begin{center}
\resizebox{\textwidth}{!}{
    \begin{tabular}{l l r | l l r}
    \toprule
    {\bf Language} & {\bf Corpus Name} & {\bf \# Utterances} & {\bf Language } & {\bf Corpus Name } & {\bf \# Utterances} \\
    \midrule
    English & TED & 268k & Mandarin & Hkust & 197k \\
    English & Switchboard & 251k & Mandarin & OpenSLR 18 & 13k \\
    English & Librispeech & 281k & Mandarin & LDC98S73  & 36k \\
    Amharic & OpenSLR 25 & 10k & Bengali & OpenSLR 37 & 196k \\
    Cebuano & IARPA-babel301b-v2.0b & 43k & Dutch & Voxforge & 8k \\
    Italian & Voxforge & 10k & Javanese & OpenSLR35 & 185k \\
    Kazakh & IARPA-babel302b-v1.0a & 48k & Kurmanji & IARPA-babel205b-v1.0a & 46k\\
    Lao & IARPA-babel203b-v3.1a & 66k & Turkish & IARPA-babel105b-v0.4 & 82k \\
    Sinhala & openSLR52 & 185k \\
    \midrule
    German & Voxforge & 41k & Mongolian & IARPA-babel401b-v2.0b & 45k \\
    Russian & Voxforge & 8k & Spanish & Callhome Hub4 & 31k \\
    Swahili & OpenSLR 25 & 10k & Tagalog & IARPA-babel106b-v0.2g  & 93k \\
    Zulu & IARPA-babel206b-v0.1e & 60k \\
    \bottomrule
    \end{tabular}
    }
    \caption{Corpora of the training set and the test set used in the experiment. Both baseline model and proposed model are trained with 17 corpus across 13 languages, and tested on 7 corpus in 7 languages. } \label{tab:corpus}
\end{center}
\end{table*}

Zero-shot learning has rarely been applied to speech sequence prediction problems. Zero-shot translation is an example of applying zero-shot learning to a different type of sequence problems\cite{johnson2017google}. In the standard settings, the zero-shot translation means that the target language pair is not in the training dataset. However, both languages should be already seen in other training pairs. In contrast, we assume a harder problem here: there is no available training audio or text for the target language at all.

In this section we describe a novel sequence model architecture for zero-shot learning. We adapt a modified ESZSL architecture from \cite{romera2015embarrassingly}. While the original architecture is devised to solve the classification problem with CNN(DECAF) features, our model aims to optimize a CTC loss over a sequence model as shown in Figure~\ref{fig:model}. We note our architecture is a general model, and it can also be used for other sequence prediction problems in zero-shot learning.

Given the training set $ \{ (\mathbf{x_n}, \mathbf{y_n}, \phi_n ), n=1 ... N \}$ where each input $\mathbf{x_n} \in \mathcal{X}$ is an utterance, $\phi_n$ is its language, and $\mathbf{y_n} \in \mathcal{Y}$ is the corresponding phoneme transcription. Suppose that $\mathbf{x_n} = (x_n^1, ..., x_n^T)$ is the input sequence where $x_n^t$ is the frame of time step $t$, and $T$ is the length of $\mathbf{x_n}$.
Each frame $x_n^t$ is first projected into a feature vector $h_n^t \in \mathbb{R}^d$ in the hidden space $\mathcal{H}$ with a Bidirectional LSTM model.
\begin{align}
                h_n^t = \theta(x_n^t; W_{\mathrm{LSTM}})
\end{align}
where $W_{\mathrm{LSTM}}$ is the parameter of the Bidirectional LSTM model. We assume that our phoneme inventory of $\phi_n$ consists of $z$ phonemes in the training set, each of them having a signature of $a$ attributes constructed as mentioned above. We can first represent our attributes in a constant signature matrix $S \in \{0,1\}^{z \times a}$ of $\phi_n$. The $(i,j)$ cell in the signature matrix is $1$ if the $i$-th phoneme has been assigned the $j$-th attribute, otherwise it is assigned to $0$. We note that while the signature matrix is constructed automatically in this work, it can be refined by linguists using phonology in each language. Then, we transform $h_n^t$ into articulatory logits with the transformation matrix $V \in \mathbb{R}^{a \times d}$. Then it is further processed into the phoneme logits $l_n^t$ with $S$.
\begin{align}
                l_n^t = S V h_n^t
\end{align}
The logits $\mathbf{l_n} = (l_n^1, ..., l_n^T)$ are then combined with $\mathbf{y_n}$ to compute the CTC loss~\cite{graves2006connectionist}. Additionally, regularizing $V$ has been proved to be useful in the original ESZSL architecture~\cite{romera2015embarrassingly}. Eventually our target is to minimize the following loss function:
\begin{align} \label{eq:s-softmax}
\displaystyle{\minimize_{V,W_{\mathrm{LSTM}}} \mathrm{CTC}(\mathbf{x_n},\mathbf{y_n}; V, W_{\mathrm{LSTM}}) + \Omega(V)}
\end{align}
where $\Omega(V)$ is an simple $\ell^2$ regularization. This objective can be easily optimized using standard gradient descent methods.

At the inference stage, we usually consider a new language $\phi_{test}$ with a new phoneme inventory. Suppose that the new inventory is composed of $z'$ phonemes, then we can automatically create a new signature matrix $S' \in \{0,1\}^{z'\times a}$, and estimate probability distribution of each phoneme $P_{acoustic}(p|x^t_n)$ from logits using $S'$ instead of $S$.


\section{Experiments}

\begin{table*}[t]
\begin{center}
\resizebox{\textwidth}{!}{
    \begin{tabular}{c c | c c c c c}
    \toprule 
    {\bf Language} & {\bf \# unseen phoneme}& {\bf Baseline PER\% } & {\bf UPM PER\%}  & {\bf Baseline Substitution\%} & {\bf UPM Substitution\%} \\
    \midrule
    German    & 2 & 68.0 & 64.9 & 51.9 & 46.9\\
    Mongolian & 18 & 87.8 & 77.5 & 44.1 & 35.8 \\
    Russian    & 19& 74.5 & 54.4 & 63.5 & 34.5 \\
    Swahili   & 2 & 55.7 & 48.9 & 27.4 & 26.6 \\
    Tagalog    & 0& 60.7 & 57.0 & 27.2 & 20.1 \\
    Spanish    & 2 & 48.6 & 44.4 & 31.0 & 26.2\\
    Zulu      & 8 & 73.1 & 67.9 & 36.2 & 33.5 \\
    \midrule
    Average  & 7.3 & 66.9 & \textbf{59.2} & 40.2 & \textbf{31.9} \\
    \bottomrule
    \end{tabular}
    }
    \caption{Phoneme error rate and phoneme substitution rate of the baseline, and our approach. Our model (UPM) outperforms the baseline for all languages, by 7.7\% (absolute) in phoneme error rate, and 8.3\% in phoneme substitution error rate.} \label{tab:result} 
    \label{table:result}
\end{center}
\end{table*}

\subsection{Dataset}
We prepare two datasets for this experiment. The training set consists of 17 corpora from 13 languages, and the test set is composed of corpora from 7 different languages. They are used by both our model and the baseline described later. Details regarding each corpus and each language are provided in Table~\ref{tab:corpus}.

We briefly describe our strategy of corpus selection in the experiment. To select the training corpus, the rich-resourced languages should be taken into account firstly to make sure the acoustic model can be fully trained. Therefore, we add three English corpora and three Mandarin corpora to the training set. Additionally, we expect both the baseline and our Universal Phonemic Model should be trained to recognize a variety of phonemes from different languages. Therefore we collect a number of corpora from different language families and diverse regions. Finally, we attempt to make the acoustic model robust to various channels and speech styles. For example, TED~\cite{rousseau2012ted} is the conference style, Switchboard~\cite{godfrey1992switchboard} is the spontaneous conversation style and Librispeech is the reading style~\cite{panayotov2015librispeech}. We note that 5 percent of the entire corpus was used as the validation set. The test corpora are selected in a similar style. They are selected from a variety of languages: not only from rich-resourced languages, but also low-resourced languages with stable audio alignments and reliable g2p models.


\subsection{Experimental Settings}
We use the EESEN framework for the acoustic modeling~\cite{miao2015eesen}. All the transcripts are transcribed into phonemes with Epitran~\cite{mortensen2018epitran}. The input feature is 40 dimension high-resolution MFCCs, the encoder is a 5 layer Bidirectional LSTM model, each layer having 320 cells. The signature matrix is designed as we discussed above, different signature matrices are used for different languages. We train the acoustic model with stochastic gradient descent, using a learning rate of 0.005. In each iteration, we apply the uniform sampling~\cite{Li2019}: first randomly select a corpus from the entire training set, and then randomly choose one batch from that corpus.


Our baseline model is the multilingual acoustic model with a shared phoneme inventory. This type of architecture is one of the standard approaches in the multilingual ASR community~\cite{tong2017investigation,vu2013multilingual}. In this architecture, all languages share a common acoustic model and a single output layer. The output layer is to predict phonemes in the universal phoneme inventory shared by all the training languages. In our experiment, the inventory consists of 131 distinct phonemes from 14 training languages. To compare the baseline with the proposed model, we also use the Bidirectional LSTM model as the encoder to compute phoneme distributions $P(p|x^t_n)$. Then we decode phonemes with greedy decoding as in our approach. We use the same configuration of LSTM architecture as well as the training criterion. As we focus on phonemic transcriptions in this work, we use phoneme error rate (PER) as the metric for evaluation. 
 

\subsection{Results}
Our results are summarized in Table~\ref{table:result}. As is shown, our approach consistently outperforms the baseline in terms of phoneme error rate. For example, the baseline achieves 55.7\% phoneme error rate when evaluated with Swahili, and our approach obtains 48.9\% in the same test set. For each language in our evaluation, we observe that we improve the phoneme error rate from 3.1\% (German) to 20.1\% (Russian) respectively. On average, the baseline has 66.9\%, and our model gets 7.7 \% better phoneme error rate.

The table also indicates the strong correlation between the number of unseen phonemes and the improvement in the phoneme error rate.  For example, Russian achieves the largest improvement with our UPM: it improves significantly by 20.1\% phoneme error rate. In our experiment, the Russian phoneme inventory has 48 phonemes in total out of which 19 of them are unseen during training. This suggests our model has a good generalization ability to adapt to languages whose acoustic contexts are rarely known.
On the other hand, every phoneme in the Tagalog inventory has been covered by other languages in the training set. Therefore, the number of its unseen phoneme is 0 and the corresponding 3.7\% phoneme error rate improvement is relatively limited. Similarly, the least improved language is German, which improved from 68.0\% to 64.9\% because there are only two unseen phonemes in German. This fact can also be explained by the relationship between German and English. German comes under the West Germanic branch in the Indo-European language family like English. As English is the largest training set in this experiment, phonemes of English are well-trained in the baseline and should be generalizing well to German. Therefore it is hard for UPM to outperform by a large margin. Additionally, we find that the correlation between the number of unseen phonemes and phoneme error rates is relatively weak. For example, Tagalog has 12\% higher phoneme error rate compared with Spanish, even its unseen phonemes are less than Spanish. This might be explained by the discrepancy of the phoneme distribution between the target language and training languages. For example, even though in principle all the phonemes of Tagalog have been covered in the training languages, their relative frequencies are not similar, which would affect the quality of the results.


To further investigate the reason for improvements for our model, we computed the (phoneme) substitution error rate, shown in the two right columns of Table~\ref{table:result}. It goes down from 40.2\% in the baseline to 31.9\% in our model. The numbers show that we have 8.3\% improvement in substitution error rate. This result suggests that our model is good at improving confusions between phonemes. However, it also indicates that our model is not able to improve addition and deletion errors.


\begin{table*}[t]
\begin{center}
\resizebox{\textwidth}{!}{
    \begin{tabular}{c c c c c}
    \toprule 
    {\bf Language} & {\bf Baseline unseen PER\%}& {\bf UPM unseen PER\% } & {\bf Baseline seen PER\%}  & {\bf UPM seen PER\%}  \\
    \midrule
    German    & 100.0 & 100.0 & 63.9 & 61.9\\
    Mongolian & 100.0 & 91.9  & 86.8 & 78.6 \\
    Russian   & 100.0 & 96.1  & 69.5 & 51.7 \\
    Swahili   & 100.0 & 86.4  & 54.3 & 46.2 \\
    Tagalog   & N.A.  & N.A.  & 57.4 & 54.2 \\
    Spanish   & 100.0 & 58.0  & 45.2 & 41.7\\
    Zulu      & 100.0 & 88.3  & 70.5 & 64.6 \\
    \midrule
    Average & 100.0 & \textbf{89.8} & 64.2 & \textbf{57.0} \\
    \bottomrule
    \end{tabular}
    }
    \caption{Phoneme error rate (\%PER) of the seen phonemes and unseen phonemes in the baseline and our approach.} 
    \label{table:unseen}
\end{center}
\end{table*}

To understand how the number of training languages contributes to the performance in the experiment, we train different models by changing the numbers of training languages: we train those models with 2, 6, 10, 14 languages. The first two languages are English and Mandarin which are corresponding to the 6 well resourced corpus in Table.\ref{tab:corpus}. The other 4, 8, 12 languages are randomly selected from the remaining training languages.

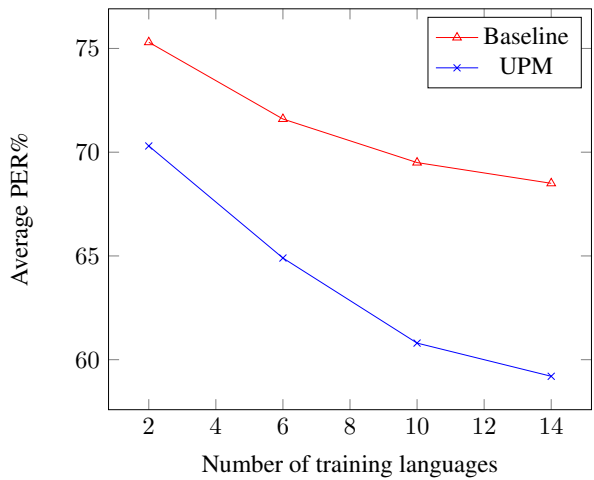
\begin{figure}[h]
\resizebox {0.45\textwidth} {!} {

\begin{tikzpicture}
	\begin{axis}[
		xlabel=Number of training languages,
		ylabel=Average PER\% ]
	\addplot[color=red,mark=triangle] coordinates {
	    (2,  75.3)
	    (6,  71.6)
	    (10, 69.5)
	    (14, 68.5)
	};
	
	\addplot[color=blue,mark=x] coordinates {
		(2,  70.3)
		(6,  64.9)
		(10, 60.8)
		(14, 59.2) 
	};
	\legend{Baseline, UPM}

	\end{axis}
\end{tikzpicture}
}
  \caption{Illustration of the relationship between the number of training languages and the average phoneme error rate over 7 languages }
  \label{fig:number}

\end{figure}

Figure.\ref{fig:number} demonstrates their performance: the red line (with triangular mark) and blue line (with cross mark) indicate the average PER of the baseline and UPM respectively. They suggest that increasing the number of training languages is helpful to reduce phoneme error rate for both models. For the baseline model, it indicates that the acoustic model get exposed to more diverse phonemes present in different languages. Therefore it learns to predict them with reduced error rates in the test set. Our UPM also improves by learning various acoustic contexts of broader articulatory attributes. The curves in Figure.\ref{fig:number} show that UPM outperforms the baseline consistently with different training size. Additionally, the gap of phoneme error rate between the two models has increased when using more languages: the gap increased from 5.0 to 9.3. The results illustrate that our UPM is better at taking advantage of the diverse training languages. Our model can infer correlations between phonemes by using their shared articulatory attributes. This ability is helpful when a specific phoneme is rarely seen but its attributes have already been well-trained using other related phonemes. On the contrary, the baseline is not adapted well to those rare phonemes or unseen phonemes. It fails to predict those phonemes when their training data are limited.

Finally, to highlight the ability of our model, we compute the phoneme error rate for each phoneme, then classify them into the seen group and unseen group based on whether the phoneme is available in the training set. To compute phoneme error rate in this case, we align the expected phonemes with the predicted phonemes using their edit distance, the phoneme error rate here denotes the correction rate for each expected phone. Table.\ref{table:unseen} demonstrates the results of both the baseline and UPM, it suggests the UPM outperforms the baseline on both groups. On average, UPM would predict 10.2 \% better for the unseen groups and 7.2 \% better for the seen groups. The average numbers demonstrate that our approach has the ability to predict unseen phonemes and could even be adapted better to seen groups. The table also shows the difficulty of the task and the weakness of our approach: we could not predict any unseen phonemes for German. The two unseen phonemes of German are /pf/ and /C/, but the frequencies of both phonemes are less than 0.5 \% in the test set, which makes the model extremely unstable when predicting those phonemes. On the other hand, the Spanish improvement of unseen PER is extremely significant, which can also be explained by the unstable prediction over low frequency unseen phonemes. Additionally, the 89.8 error rate of unseen groups is still not practical in the real-world production systems. 

\section{Related Work}
\label{sec:related}
We briefly outline several areas of related works, and describe their connections and differences with this paper. Zero-shot learning was first applied to recognize unseen objects during training in the computer vision field~\cite{lampert2009learning,palatucci2009zero,socher2013zero}. However those works rarely mention speech recognition.


Meanwhile there has been growing interests in zero-resource speech processing \cite{glass2012towards,jansen2013summary}, most of the work focusing on tasks like acoustic unit discovery, unsupervised segmentation and spoken term discovery~\cite{heck2017feature}. These models are useful for various extrinsic speech processing tasks like topic identification. However, the unsupervised concept cannot be directly grounded to actual phonemes, hence making it impracticable to do speech recognition or acoustic modeling. The usual intrinsic evaluations that these zero resource tasks are tested on is ABX discriminability task or the unsupervised word error rate which are good for quality estimates but not practical as they use an oracle or ground truth labels to assign cluster labels. In addition these approaches demands a modest size of audio corpus of targeting language (e.g: 2.5h to 40h). In contrast, our approach assumes no audio corpus and no text corpus for targeting languages. The idea of decomposing speech into concepts was also discussed by \cite{lake2014one}, where the authors propose a generative model to learn representations for spoken words which they then use to classify words with only one training sample available per word. Though this is in the same line as the zero-resource speech processing papers, we feel the motivation behind the decomposition is very similar to this work. 

Another group of researchers explore adaptation techniques for multilingual speech recognition, especially for low resource languages. In these multilingual settings, the hidden layers are either HMM or DNN models which are shared by multiple languages, and the output layer is either language specific phone set or a universal IPA-based phone set \cite{tong2017investigation,vu2013multilingual,thomas2010cross,chen2015multitask,dalmia2018sequence}. However predictable phonemes are restricted to the phonemes in the training set, thus they fail to predict unseen phonemes in the test set. In contrast, our model can predict unseen phonemes by taking advantage of their articulatory attributes. 

Articulatory features have been shown to be useful in speech recognition under several situation. Specifically, articulatory features has been used to improve robustness under noisy and reverberant environment~\cite{kirchhoff1998combining}, compensate for crosslingual variability~\cite{stuker2003multilingual},  improve word error rate in multilingual models~\cite{stuker2003integrating}, be beneficial for low resource languages~\cite{muller2016towards}, detecting spoken words~\cite{prabhavalkar2013discriminative}, clustering phoneme-like units for unwritten languages~\cite{muller2017improving}, recognizing unseen languages~\cite{siniscalchi2011experiments}, developing phonological vocoder~\cite{cernak2016phonvoc}. There are also some attempts to predict articulatory features or distributions for clinical usages~\cite{jiao2017interpretable,vasquez2019phonet}, but they do not provide a model to predict unseen phonemes. 

We note that there are also several attempts to build acoustic models for unseen phonemes. For example, the authors in \cite{scharenborg2018building} present an interesting method to predict unseen phonemes in Mboshi by mapping Dutch/Mboshi phonemes in the same space using an extrapolation approach. However starting phonemes used for extrapolation had to be manually assigned for every missing phoneme and every pair of languages. Compared with this work, our model proposes a much more generic algorithm to recognize unseen phonemes. Another previous work integrated articulatory attributes into the state-position based decision tree to predict unseen phones in their multilingual model~\cite{knill2014language}, however the approach is limited to traditional HMM models and it is unclear how attributes are extracted and how it performs when predicting unseen phonemes. 



\section{Conclusion}
In this work, we propose the Universal Phonemic Model to apply zero-shot learning to the automatic phonemic transcription task. Our experiment shows that it outperforms the baseline by 7.7\,\% phoneme error rate on average for 7 languages. While the performance of our approach is still not enough for the real-world production systems, it paves the way to tackle zero-shot learning of speech recognition with a new framework. 

\section{Acknowledgements}
This project was sponsored by the Defense Advanced Research Projects Agency (DARPA) Information Innovation Office (I2O), program: Low Resource Languages for Emergent Incidents (LORELEI), issued by DARPA/I2O under Contract No. HR0011-15-C-0114. 

\bibliography{aaai}
\bibliographystyle{aaai}

\end{document}